\documentclass[sigconf]{acmart}
\usepackage{lipsum} 
\AtBeginDocument{%
  \providecommand\BibTeX{{%
    \normalfont B\kern-0.5em{\scshape i\kern-0.25em b}\kern-0.8em\TeX}}}
 
\acmConference[DSHealth at SIGKDD '22]{Workshop on Applied Data Science for Healthcare, ACM SIGKDD Conference on Knowledge Discovery and Data Mining}{August 14-18, 2022}{Washington, D.C.}
\usepackage{subcaption}

\begin{document}

\title{Using Interpretable Machine Learning to Predict Maternal and Fetal Outcomes}

\author{Tomas M. Bosschieter}
\email{tomasbos@stanford.edu}
\affiliation{%
  \institution{Stanford University}
  \streetaddress{475 Via Ortega}
  \city{Stanford}
  \state{CA}
  \country{USA}
  \postcode{94305}
}

\author{Zifei Xu}
\email{zifei98@stanford.edu}
\affiliation{%
  \institution{Stanford University}
  \streetaddress{475 Via Ortega}
  \city{Stanford}
  \state{CA}
  \country{USA}
  \postcode{94305}
}

\author{Hui Lan}
\email{huilan@stanford.edu}
\affiliation{%
  \institution{Stanford University}
  \streetaddress{475 Via Ortega}
  \city{Stanford}
  \state{CA}
  \country{USA}
  \postcode{94305}
}

\author{Benjamin J. Lengerich}
\email{blengeri@mit.edu}
\affiliation{%
  \institution{Massachusetts Institute of Technology}
  \city{Cambridge}
  \state{MA}
  \country{USA}
}

\author{Harsha Nori}
\email{hanori@microsoft.com}
\affiliation{%
  \institution{Microsoft Research}
  \city{Redmond}
  \state{WA}
  \country{USA}
}

\author{Kristin Sitcov}
\email{ksitcov@qualityhealth.org}
\affiliation{%
  \institution{Foundation for Health Care Quality}
  \city{Seattle}
  \state{WA}
  \country{USA}
}

\author{Vivienne Souter, MD}
\email{vsouter@qualityhealth.org}
\affiliation{%
  \institution{Foundation for Health Care Quality}
  \city{Seattle}
  \state{WA}
  \country{USA}
}

\author{Rich Caruana}
\email{rcaruana@microsoft.com}
\affiliation{%
  \institution{Microsoft Research}
  \city{Redmond}
  \state{WA}
  \country{USA}
}

\renewcommand{\shortauthors}{Bosschieter, et al.}

\begin{abstract}
  Most pregnancies and births result in a good outcome, but complications are not uncommon and when they do occur, they can be associated with serious implications for mothers and babies. Predictive modeling has the potential to improve outcomes through better understanding of risk factors, heightened surveillance, and more timely and appropriate interventions, thereby helping obstetricians deliver better care. For three types of complications we identify and study the most important risk factors using Explainable Boosting Machine (EBM), a glass box model, in order to gain intelligibility: (i) Severe Maternal Morbidity (SMM), (ii) shoulder dystocia, and (iii) preterm preeclampsia. While using the interpretability of EBM's to reveal surprising insights into the features contributing to risk, our experiments show EBMs match the accuracy of other black-box ML methods such as deep neural nets and random forests. 
\end{abstract}

\ccsdesc[500]{Applied computing~Health informatics}
\ccsdesc[300]{Computing methodologies~Machine learning}
\keywords{Explainability, intelligible models, generalized additive models, AI for healthcare, obstetrics}


\maketitle

\section{Introduction}
Of the 3.6 million births per year in the U.S. \cite{osterman2022births}, Severe Maternal Morbidity (SMM) happens in as many as 60,000 cases \cite{declercq2020maternal}, leading to serious short- or long-term consequences for the mother's health. In our study, SMM is a collective term for 6 adverse diagnoses: (i) hysterectomy, (ii) blood transfusion, (iii) disseminated intravascular coagulation, (iv) amniotic fluid embolism, (v) thromboembolism, and (vi) eclampsia.  Additionally, shoulder dystocia and (preterm) preeclampsia are two other common serious conditions, putting the baby's and mother's lives at risk. If there were a timely risk indication, many cases of SMM, shoulder dystocia and (preterm) preeclampsia could potentially be avoided \cite{creanga2014racial, declercq2020maternal}. While several analyses and black-box models have been deployed to predict SMM and its risk factors \cite{gao2019learning, cartus2021can, callaghan2008identification}, preeclampsia \cite{bennett2022imbalance, moreira2017predicting, jhee2019prediction}, and shoulder dystocia \cite{tsur2020development, bartal2021651}, the risk factors for each have remained severely under-studied. Additionally, models trained on international data (e.g. \cite{squires2011us}) might not perform as well in U.S.-based hospitals \cite{yamada2021external}. 

Equipped with recent clinical data containing 408 features and 222,001 de-identified patients, we use interpretable machine learning models -- Explainable Boosting Machines (EBMs) -- to uncover the most important risk factors. We show that EBMs yield an Area Under the Receiver Operating Characteristic curve (AUROC) on par with XGBoost \cite{chen2016xgboost}, random forests \cite{breiman2001random}, deep neural networks (DNNs) \cite{gardner1998artificial}, and logistic regression \cite{kleinbaum2002logistic} while remaining fully interpretable.

Our main contributions are:
\begin{itemize}
    \item Showcasing how intelligible models reveal surprising risk contribution relationships not traditionally recognized.
    \item Illustrating EBM's potential for healthcare applications, rivalling the current industry-standard in obstetrics, logistic regression \cite{LR}, by outperforming it while also providing intelligibility.
    \item Leveraging a new and robust data set in healthcare.
\end{itemize}

\section{Preliminaries}\label{sec:prelims}
For a target variable $Y$ and predictor features $x_1, \ldots, x_n$, Generalized Additive Models (GAMs) \cite{hastie2017generalized} generalize linear (regression) models $Y = b_0 + a_1x_1 + \cdots + a_nx_n$ to an additive model with univariate shape functions $f_i$ and a situation-dependent link function $g$ (e.g. logistic for classification and identity for regression):
\begin{equation}
    g(\mathbb{E}[Y]) = \beta_0 + f_1(x_1) + f_2(x_2) + \cdots + f_n(x_n)
\end{equation}
Explainable Boosting Machines emerge when the $f_i$ are learned by boosting shallow decision trees in a round-robin fashion \cite{lou2012intelligible, nori2019interpretml}. Additionally, EBMs can be trained to automatically detect important pairwise interaction terms to further boost accuracy while preserving intelligibility \cite{lou2013accurate}. Because EBMs are GAMs, all decision trees for the $i$-th feature are added to generate the shape function $f_i$ and corresponding graph, which displays the i-th feature's effect on the target outcome. EBMs have recently gained popularity because of their local and global interpretability, from applications in healthcare \cite{zhou2020identifying} to predicting sports outcomes \cite{decroos2019interpretable, xenopoulos2022analyzing}, slope failures \cite{maxwell2021explainable}, and modeling dark matter \cite{cannarozzo2021merger}.

\section{Experimental Setup and Methods}\label{sec:experimental-setup}

\subsubsection{Data set}\label{sec:dataset}
The real-world, clinical data used in this analysis are from the Foundation for Health Care Quality’s Obstetrical Care Outcomes Assessment Program
(OB COAP) \cite{fhcq_obcoap}, including de-identified patient-level information from medical records at 20 hospitals collected for quality improvement purposes. Out of the 222,001 patients, 3102 (1.40\%) were SMM-positive, 4908 (2.21\%) shoulder dystocia-positive, and 4555 (2.05\%) preterm preeclampsia-positive.


\subsubsection{Data analysis}\label{sec:data-analysis}
For each outcome, we use clinical expertise to identify and select the features that can be measured before the outcome occurs. We select those features, and do not use the other features to train any of the machine learning models. For example, in trying to predict preterm preeclampsia for pregnant women\footnote{Note that predicting preterm preeclampsia for pregnant women can only be done in practice $<37$ weeks into the pregnancy, otherwise it's no longer considered preterm.}, the feature `length of time till delivery' should not be used by a machine learning model, since labor has not taken place yet and so it's not known at the time of prediction, namely < 37 weeks into the pregnancy. However, `length of time till delivery' is selected as a feature to be used in predicting SMM and shoulder dystocia.

To obtain training and test data sets, we stratify by hospital, and select a subset of hospitals such that the union of their patients represents 75\% of all patients. We use these patients for training, and the other 25\% are used as test data for validation. This is a form of external cross-validation, since some hospitals and all their patients are used for training, while a distinct other group of hospitals are used for validation. Note that this is standard practice in medical sciences \cite{kamran2022early, lin2020external}. Furthermore, we use the standard procedure of dummy encoding for categorical data, and impute missing values with the mean. No data normalization is required.

Lastly, as a preprocessing step, we discard birth events whose data are deemed not realistic by clinical experts (and assumed to be entered into the system erroneously), taking a conservative approach to avoid accidentally discarding valid data. For example, we never include birth events with a negative length of time till delivery, births where the baby seemingly weighs over 8000 grams right after labor, and we also exclude cases where the pregnant patient has a BMI over 120.

\subsection{Model parameters}\label{sec:methods}
For EBMs, we use hyperparameters \verb|outer_bags=25|, \verb|inner_bags=25|, \verb|min_samples_leaf=25|, \verb|interactions=20|. For XGBoost we use \verb|eta=0.04|, \verb|subsample=0.7|, and \verb|max_depth=5|. Random forests are trained with \verb|n_estimators=1000|, \verb|min_samples_split=60|, and \verb|min_samples_leaf=40|. Lastly, the DNN is an MLP with 7 hidden layers of 200 neurons each. The hyperparameters of all models were each determined using 5-fold Cross-Validation to maximize AUROCs while ensuring good calibration and using heuristic methods to minimize overfitting; all parameters not mentioned are common defaults. We use 5-fold CV for all models to make computations feasible.

\section{Results}\label{sec:results}
\begin{figure*}[h]
    \centering
    \begin{subfigure}{.48\textwidth}
        \centering
        \includegraphics[width=1\linewidth]{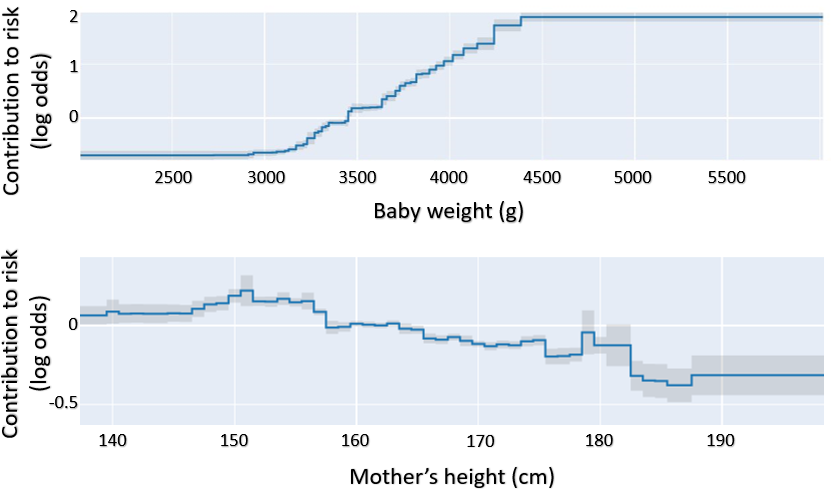}
        \caption{The individual contributions of baby weight and maternal height to the risk of shoulder dystocia.}
        \label{fig:ShD-babyweight_and_maternal_height}
    \end{subfigure}%
    \hfill
    \begin{subfigure}{.48\textwidth}
        \centering
        \includegraphics[width=.8\linewidth]{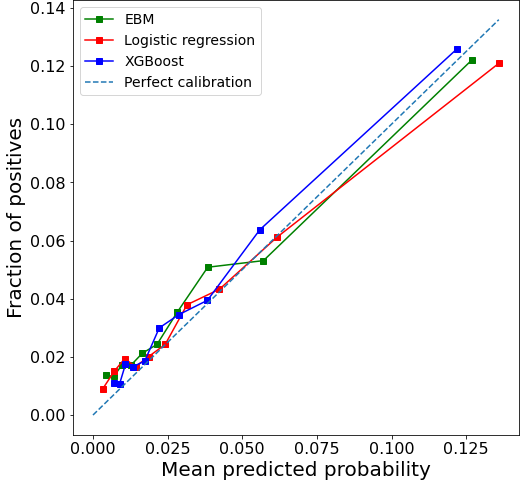}
        \caption{Calibration plot for shoulder dystocia using 10 bins.}
        \label{fig:cal-plots}
    \end{subfigure}
    \caption{Along with a calibration plot, we show two shape functions for shoulder dystocia: the risk contribution of (i) baby weight, and (ii) the mother's height. Similar calibration is achieved for other models, which are omitted to allow for detailed inspection.}
\label{fig:ShD-ind-and-calplot}
\end{figure*}

For each outcome (SMM, shoulder dystocia, preterm preeclampsia) we compare EBMs to XGBoost, random forests, deep neural networks in the form of a multilayer perceptron, and logistic regression. The Area Under the ROC curve (AUROC) for each model and outcome can be found in Table~\ref{tab:aurocs}. The calibration plot for shoulder dystocia is shown in Figure~\ref{fig:cal-plots}. We exclude the calibration plots of random forests and deep neural networks to allow for close inspection of the graph -- they achieve similar calibrations. Similar calibrations are achieved for other outcomes as well. 

\begin{table*}[!h]
    \caption{A comparison between the Area Under the ROC curve (AUROC) for each outcome. Higher is better.}
    \centering
    \begin{tabular}{c|c c c c c}
    \toprule
          Outcome & \textbf{EBM} & \textbf{Logistic Regression} & \textbf{XGBoost} & \textbf{Random Forests} & \textbf{DNN} \\
         \midrule
         SMM &$0.756\pm 0.020$ & $0.748\pm 0.022$& $0.761\pm 0.019$& $0.761\pm 0.018$ &$0.742\pm 0.016$ \\
         Shoulder dystocia  & $0.744 \pm 0.017$ & $0.744 \pm 0.020$ & $0.751 \pm 0.018$ & $0.713 \pm 0.019$ & $0.752 \pm 0.019$ \\
         Preterm Preeclampsia  &$0.770 \pm 0.006$ &$0.750 \pm 0.022$ &$0.767 \pm 0.018$ & $0.749 \pm 0.019$ & $0.758 \pm 0.014$ \\
         \midrule
         Mean AUROC & $0.757 \pm 0.014$ & $0.747 \pm 0.021$ & $0.760 \pm 0.055$ & $0.741 \pm 0.019$ & $0.751 \pm 0.016$
    \end{tabular}
    \label{tab:aurocs}
\end{table*}


\begin{figure*}[h]
    \centering
    \begin{subfigure}{.48\textwidth}
        \centering
        \includegraphics[width=1\linewidth]{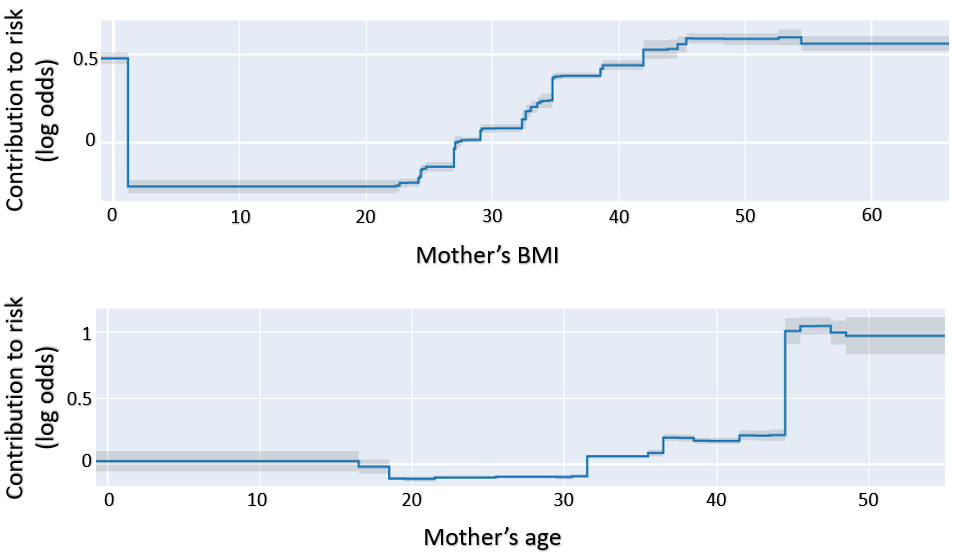}
        \caption{Displaying the individual risk contributions of the mother's BMI and the mother's age for preterm preeclampsia.}
        \label{fig:PP-mothersAge_BMI}
    \end{subfigure}%
    \hfill
    \begin{subfigure}{.48\textwidth}
        \centering
        \includegraphics[width=1\linewidth]{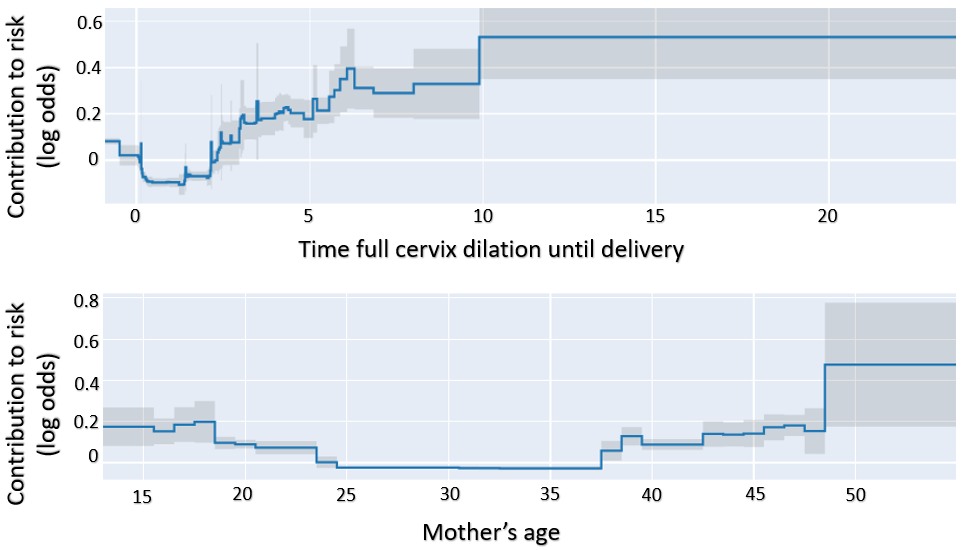}
        \caption{Risk contribution of (i) the time from full cervical dilation to delivery, and (ii) the mother's age to SMM.}
        \label{fig:SMM-2plots}
    \end{subfigure}
    \caption{Important shape functions for preterm preeclampsia (left) and SMM (right).}
\label{fig:shapefuncs-2PP-2SMM}
\end{figure*}

EBMs allow each feature's contribution to risk to be easily visualized. Figures~\ref{fig:ShD-babyweight_and_maternal_height},~\ref{fig:PP-mothersAge_BMI},~\ref{fig:SMM-2plots} show two shape functions for each outcome, Observe that the error bars shown in the shape functions in Figures~\ref{fig:ShD-babyweight_and_maternal_height},~\ref{fig:PP-mothersAge_BMI},~\ref{fig:SMM-2plots} represent the standard deviations yielded by the outer-bagging process employed by EBMs.

\begin{figure*}[h]
    \centering
    \begin{subfigure}{.48\textwidth}
        \centering
        \includegraphics[width=1\linewidth]{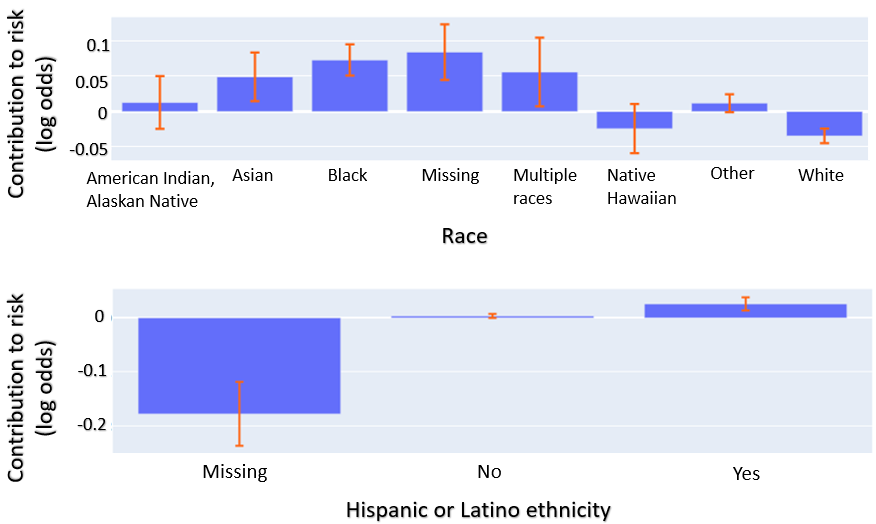}
        \caption{Shoulder dystocia.}
        \label{fig:ShD-race-ethn}
    \end{subfigure}%
    \hfill
    \begin{subfigure}{.48\textwidth}
        \centering
        \includegraphics[width=1\linewidth]{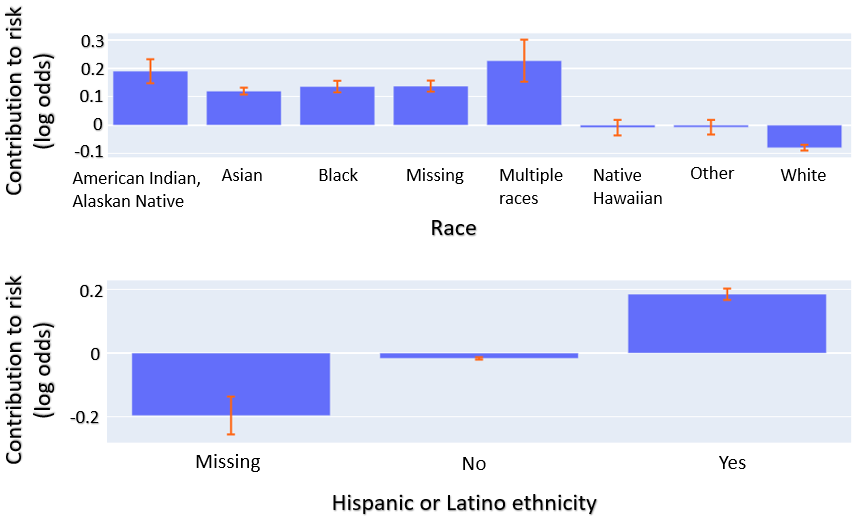}
        \caption{SMM.}
        \label{fig:SMM-race-ethn}
    \end{subfigure}
    \caption{The risk contribution of race and ethnicity for shoulder dystocia and SMM. Preterm preeclampsia is similar. For SMM, the `Missing' bucket represents 5.6\% of the data for the feature `Race', and 3.7\% for `Hispanic or Latino ethnicity'. For shoulder dystocia, these are 4.8\% and 4.0\% respectively.}
\label{fig:race-ethn}
\end{figure*}

Furthermore, Figure~\ref{fig:race-ethn} shows the risk contribution of race and ethnicity for shoulder dystocia and SMM. We see a clinically surprising near-linear relationship between baby weight and risk in Figure~\ref{fig:ShD-babyweight_and_maternal_height} for weights between 3250g and 4250g. Using the metric of the mean absolute contribution to log-odds, the most important risk factors for each outcome are ordered in Table~\ref{tab:most-imp-feats}.

\begin{table*}[h]
    \caption{The three most important contributors to the risk for each outcome according to the EBM.}
    \centering
    \begin{tabular}{c|c|c|c}
    \toprule
          Rank & \textbf{SMM} & \textbf{Shoulder dystocia} & \textbf{Preterm preeclampsia} \\
         \midrule
         1 & Preeclampsia/gestational hypertension & Baby's weight & Mother's BMI \\
         2 & Time full cervical dilation to delivery & Time membranes rupture till delivery & Num. previous Stillbirths \\
         3 & C-section & Mother's height & Pre-pregnancy hypertension
    \end{tabular}
    \label{tab:most-imp-feats}
\end{table*}

\section{Discussion}\label{sec:discussion}

As shown in Table~\ref{tab:aurocs} and Figure~\ref{fig:cal-plots}, we see that EBMs are well-calibrated models with an AUROC on par with other models. Surprisingly, EBMs find that the biggest contributors to risk aren't necessarily the traditionally recognized ones. The risk factors usually associated with shoulder dystocia are diabetes and baby size, while the EBM also shows that time from rupture of membranes to delivery, and the mother's height are of similar importance, while reiterating the great influence of the baby's weight. Similarly, the main indicators picked up by the EBM for SMM are preeclampsia/gestational hypertension, time from full cervical dilation to delivery, and c-section, while baby weight and number of weeks on delivery also play major roles, instead of the mother's maternal age, race and ethnicity. Note, however, that the effects of one's race are often embedded in other correlated factors like the level of accessible care. Lastly, the largest contributions to the risk of preterm preeclampsia are made by the mother's BMI, number of previous stillbirths, and pre-pregnancy hypertension, while the mother's age also plays a major role, especially for women older than 44; see Figure~\ref{fig:PP-mothersAge_BMI}. The comparison to SMM regarding the risk contribution of the mother's age is noteworthy; see Figure~\ref{fig:SMM-2plots}. The contribution of race and ethnicity is shown in Figure~\ref{fig:race-ethn}. Both figures suggest non-White race and Hispanic/Latino ethnicity are associated with higher risks for both shoulder dystocia and SMM.

The perceived jumps in Figure~\ref{fig:ShD-babyweight_and_maternal_height}, and other contribution graphs as well, are likely due to clinical interventions happening at clinically relevant cut-offs. Figure~\ref{fig:ShD-babyweight_and_maternal_height} uses actual birth weight (as opposed to estimated birth weight) to highlight the importance of fetal weight for shoulder dystocia. This alludes to the value more research and improved estimates of prenatal weight could play in predicting maternal and fetal outcomes. Additionally, in light of the importance of a mother's height to the risk of shoulder dystocia, we propose including pelvimetry in clinical examinations, reiterating the relationship between the mother's height and pelvic opening.

\section{Conclusions}\label{sec:conclusions}
We leverage a new, robust data set together with the intelligibility of EBMs to illustrate EBMs' potential to improve healthcare and mitigate risk in pregnancy. We show that EBMs reach an AUROC and calibration similar to other popular methods such as XGBoost, logistic regression, and deep neural networks. Additionally, inspecting the shape functions learned by EBMs reveals surprising results that challenge  traditional beliefs about the main contributors to pregnancy risks, and suggests more research should be done to accurately estimate fetal weights.

\begin{acks}
We are deeply indebted to Professor Trevor Hastie for the intellectual discussions and his great feedback. We also thank the sites participating in OB COAP, the Foundation for Health Care Quality (FHCQ) and the OB COAP Management Committee for supporting this work. Additionally, FHCQ's Ian Painter has helped tremendously in preparing the data.
\end{acks}

\bibliographystyle{ACM-Reference-Format}
\bibliography{sample-base}






\end{document}